\documentclass[conference]{IEEEtran}
\ifCLASSINFOpdf
\else
\fi

\usepackage{times}
\usepackage{amsmath}
\usepackage{amssymb}
\usepackage{algpseudocode}
\usepackage{algorithm}
\usepackage{graphicx}
\usepackage{caption}
\usepackage{subcaption}
\usepackage{bm}
\usepackage{url}
\let\labelindent\relax
\usepackage{enumitem}
\usepackage{color}
\usepackage{multirow}

\newcommand{\K}[0]{\mathbf{K}}
\newcommand{\Pb}[0]{\mathbf{P}}
\newcommand{\Pv}[0]{\bm{\mathcal{P}}}
\DeclareMathOperator{\Tr}{Tr}
\DeclareMathOperator{\diag}{diag}
\newtheorem{mydef}{Definition}
\newtheorem{theorem}{Theorem}

\begin{document}

\title{Multi-view Unsupervised Feature Selection by Cross-diffused Matrix Alignment}

\author{
\IEEEauthorblockN{Xiaokai Wei,
Bokai Cao and
Philip S. Yu}
\IEEEauthorblockA{Department of Computer Science\\
University of Illinois at Chicago,
Chicago, IL\\ Email: \{xwei2, caobokai, psyu\}@uic.edu}
}

\maketitle
\begin{abstract}
Multi-view high-dimensional data become increasingly popular in the big data era. Feature selection is a useful technique for alleviating the curse of dimensionality in multi-view learning. In this paper, we study unsupervised feature selection for multi-view data, as class labels are usually expensive to obtain. Traditional feature selection methods are mostly designed for single-view data and cannot fully exploit the rich information from multi-view data. Existing multi-view feature selection methods are usually based on noisy cluster labels which might not preserve sufficient information from multi-view data. To better utilize multi-view information, we propose a method, CDMA-FS, to select features for each view by performing alignment on a cross diffused matrix. We formulate it as a constrained optimization problem and solve it using Quasi-Newton based method. Experiments results on four real-world datasets show that the proposed method is more effective than the state-of-the-art methods in multi-view setting.

\end{abstract}

\section{Introduction}

\begin{figure*}[t]
\includegraphics[width=0.8\textwidth]{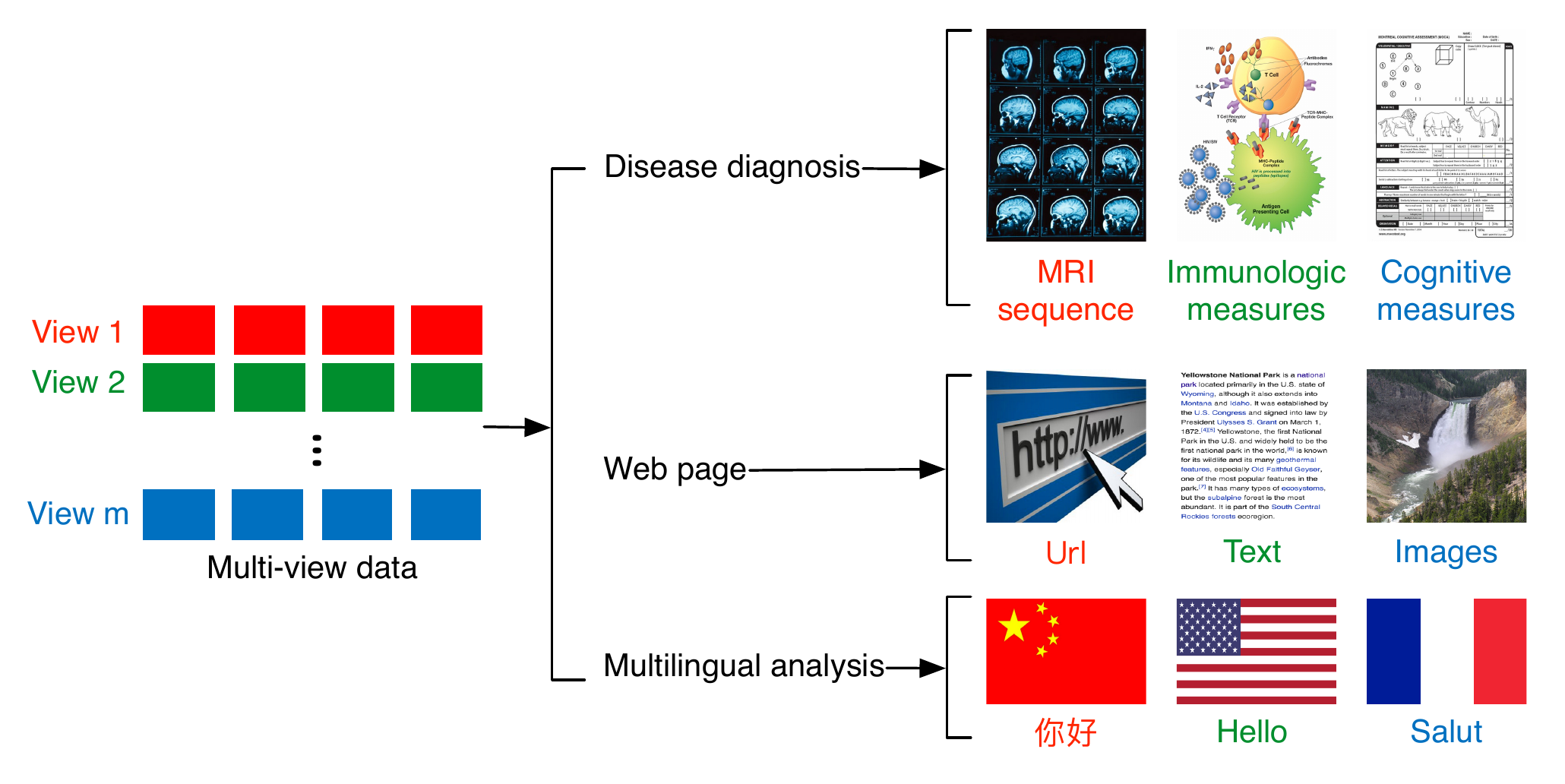}
\centering
\caption{Examples of multi-view data}
\label{fig:multiview_intro}
\end{figure*}

Data obtained from different sources or feature subsets usually provide complementary information for machine learning tasks, and conventionally they are named as multi-view data. We can observe multi-view data in a wide range of application domains (Figure \ref{fig:multiview_intro}). For example, news about the same event can often be reported in different languages and by different agencies. In the video domain, in addition to features extracted from visual signals, videos are often equipped with textual descriptions and related tags. In medical science, many different diagnosis tools have been developed to obtain a large number of measurements from various laboratory tests, including clinical, imaging, immunologic, serologic features.

Capability for simultaneous consideration of data coming from multiple views/sources is important for many learning tasks, which is referred to as multi-view learning. Multiple views together depict an enriched picture about the entities of interest and thereby provide an effective way of heterogeneous data fusion. How to effectively incorporate the abundant information from multiple views is critical for different application domains \cite{Kumar11} \cite{Tang13}. It has been shown that incorporating information from multiples views can improve the performance of various machine learning tasks. For example, co-regularized spectral clustering \cite{Kumar11}, by enforcing consensus learning on latent factors, outperforms single-view clustering significantly.

The curse of dimensionality is an inevitable problem in the era of big data, which is also one of the major challenges in many multi-view learning scenarios. For example, the vocabulary of news articles can contain more than $100,000$ words in each language. Also, the user generated content in social media (such as blog websites) tends to be highly noisy. Such high-dimensional noisy data can hamper the performance and efficiency of many machine learning/data mining tasks. Feature selection is potentially a useful technique for alleviating such issue. Traditional feature selection methods mainly focus on a single view which could be insufficient considering the existence of other views being available. It is desirable to utilize information from other complementary views, when selecting features for each view.

\begin{figure}[t!]
\centering
    \begin{subfigure}[t]{0.4\textwidth}
        \centering
        \includegraphics[height=2.0in]{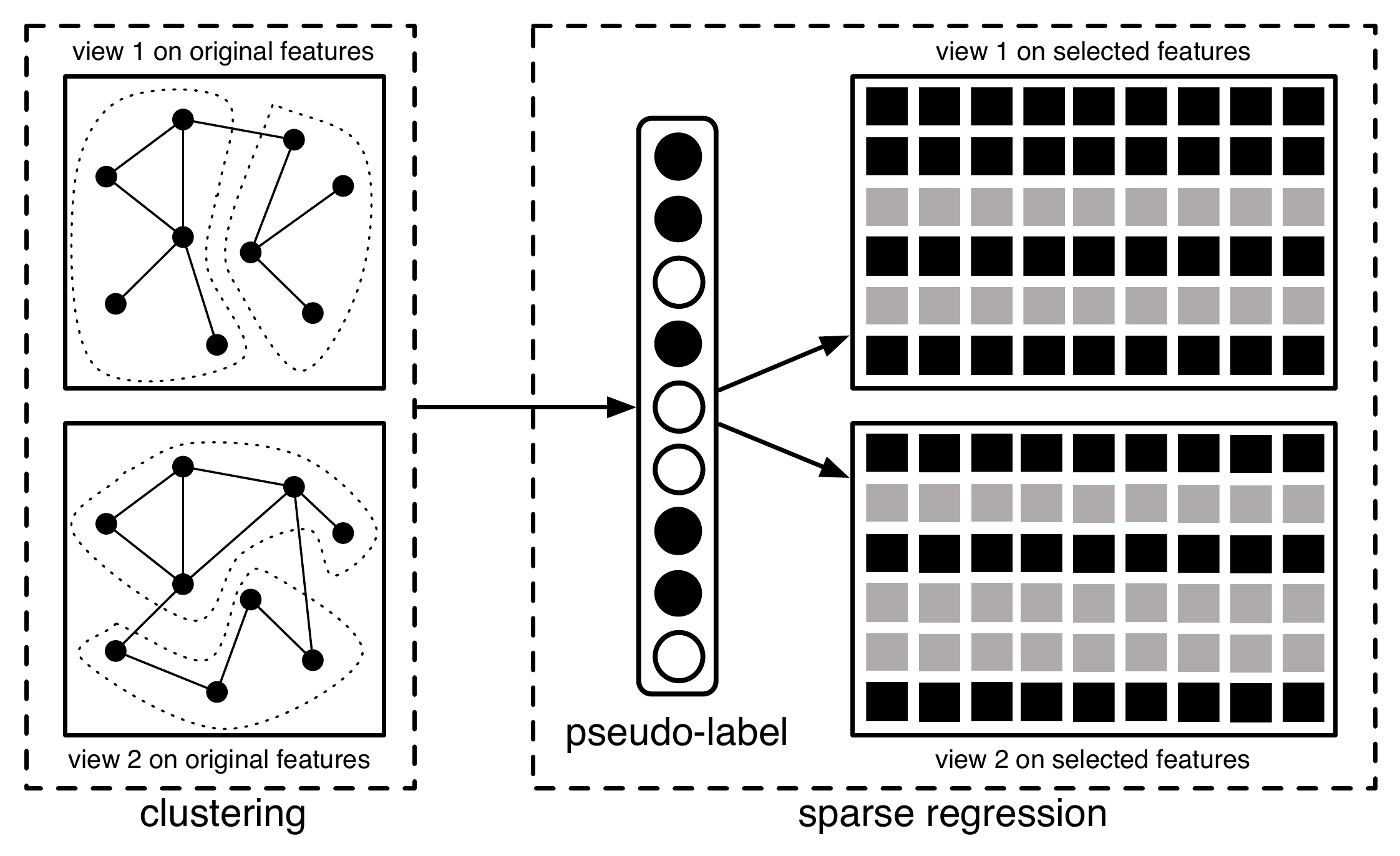}
        \caption{Pseudo-label based multi-view feature selection}
    \end{subfigure}%
    
    \begin{subfigure}[t]{0.4\textwidth}
        \centering
        \includegraphics[height=2.0in]{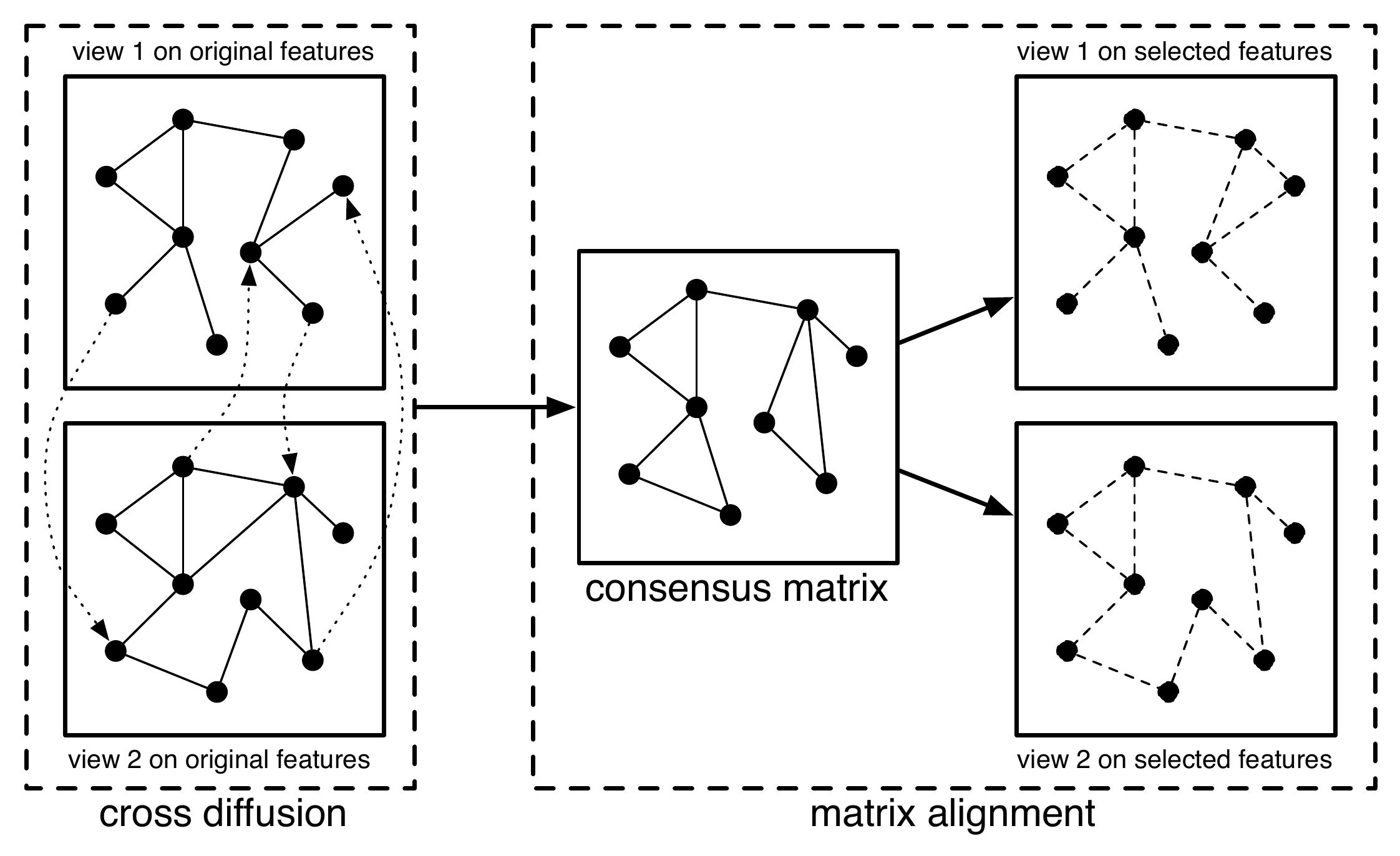}
        \caption{Multi-view feature selection by cross-diffused matrix alignment}
    \end{subfigure}
\caption{Comparison of CDMA-FS framework with existing multi-view feature selection methods. Lines between black dots represent the similarity relationship between data instances.}
\label{fig:cdma_fs}
\end{figure}

Since class labels are usually expensive to obtain, unsupervised feature selection usually has wider applicability than its supervised counterpart. The key challenge of unsupervised multi-view feature selection is twofold: (1) how to effectively represent the fused information from multiple views, and (2) how to effectively exploit the fused information representation to select high-quality features. State-of-the-art unsupervised multi-view feature selection approaches \cite{Tang13} \cite{Qian14} fuse information by generating intermediate cluster labels. However, summarizing the information for each instance with a cluster label tends to lose too much information, since the cluster labels are usually noisy and inaccurate. In this paper, we propose a new method, CDMA-FS (Cross Diffused Matrix Alignment based Feature Selection), to address the challenges of multi-view feature selection in unsupervised setting. The advantages of our method compared to state-of-the-art approaches \cite{Tang13} \cite{Qian14} can be summarized as follows.
\begin{itemize}
    \item We employ a cross diffusion-based approach to learn a consensus similarity graph from multiple views, which retains more information than the cluster labels (Figure \ref{fig:cdma_fs}).
    \item Rather than relying on cluster-label guided sparse regression, we directly exploit the information from the cross-diffused matrix by matrix alignment.
    \item Existing approaches typically have a few parameters which are difficult to set in unsupervised setting. This makes them less practical for real-world applications. In contrast, we provide guidelines for setting the parameter in the proposed method.
    \item Our objective function is not based on linear regression and hence can evaluate the non-linear usefulness of features. 
\end{itemize}

The rest of paper is organized as follows. In section II, we review related work on unsupervised feature selection on single-view or multi-view data. In section III $\sim$ V, we present the approach of CDMA-FS by aligning with cross-diffusd matrix. Experimental results are discussed in section VI and we conclude our work in section VII.

\section{Related Work}

Earlier unsupervised feature selection methods \cite{He05} \cite{Zhao07} usually assign scores to each feature based on certain heuristics and neglect the correlation among features. However, such heuristic based methods usually ignore the correlation among the features and redundancy may exist in the selected features. In recent years, different methods \cite{Yang11} \cite{Qian13} \cite{Wei15} \cite{Wei16b} have been proposed to evaluate feature quality jointly. Linear projection based methods \cite{Yang11} \cite{Li12} \cite{Du15} \cite{Wang15} with sparsity-inducing $L_{2,1}$ norm have become prevalent among others. Compared to the heuristic-based methods \cite{He05} \cite{Zhao07}, the major advantage of $L_{2,1}$-based approaches is that they can evaluate features jointly. Different $L_{2, 1}$ norm-based methods usually differ in the ways they generate pseudo labels and the loss functions on the projection.  Unsupervised Discriminative Feature Selection (UDFS) \cite{Yang11} introduces pseudo-label based regression to better capture the information from the local structure. Non-negative Discriminative Feature Selection (NDFS) \cite{Li12} derives the cluster/pseudo labels from non-negative spectral analysis. Robust Unsupervised Feature Selection (RUFS) \cite{Qian13} and Embedded Unsupervised Feature Selection (EUFS) \cite{Wang15} generate pseudo labels from non-negative matrix factorization. Robust Spectral Feature Selection (RSFS) \cite{Shi14} employs local kernel regression for the cluster indicators and Huber loss for the projection. These methods are only able to evaluate the. To address this issue, Stochastic Neighbor-preserving Feature Selection (SNFS) \cite{Wei16a} and Nonlinear Joint Feature Selection (NJFS) \cite{Wei16} are proposed, which can evaluate the non-linear usefulness of features.


Recently, several pseudo label-based methods have been extended to multi-view setting \cite{Tang13} \cite{Qian14} \cite{Shao16} via cluster consensus learning. In these approaches, pseudo-labels derived from certain clustering algorithms are required to be the same across different views in order to incorporate multi-view information. For example, adaptive Unsupervised Multi-view Feature Selection (AUMFS) \cite{Feng12} rely on spectral clustering on the combined similarity graphs obtained from different views. Multi-View Feature Selection (MVFS) \cite{Tang13} and MVUFS \cite{Qian14} can be seen as extention of NDFS \cite{Li12} and RUFS \cite{Qian13} to multi-view feature selection by enforcing consensus on the cluster indicators from different views, respectively. However, they rely on the cluster labels to guide feature selection, and the noisy cluster labels may lead to suboptimal feature selection results. Also, they evaluate features based on linear regression and hence cannot select high-quality features if they are non-linearly correlated with the class labels.

\section{Fusing Different Views by Cross Diffusion}

We denote $n$ data samples with $m$ views as $\{ \mathbf{X}^{(v)}|v = 1, \dots, m\}$, $\mathbf{X}^{(v)}=[\mathbf{x}_1^{(v)}, \mathbf{x}_2^{(v)}, \dots, \mathbf{x}_n^{(v)}]$ and the number of features in the $v$-th view as $D^{(v)}$. So $\mathbf{x}_i^{(v)} \in \mathbb{R}^{D^{(v)}}$ and $x_{ip}^{(v)}$ denotes the value of $p$-th ($p = 1, \dots, D^{(v)}$) feature of $\mathbf{x}_i^{(v)}$. 

The proposed CDMA-FS framework is a two-step approach. First, we fuse different kernels into one robust similarity matrix through cross diffusion. Second, we perform matrix alignment for the features from each view so that the kernel constructed from the selected features can best align with the fused matrix (Figure \ref{fig:cdma_fs}). In this manner, feature selection on each view can benefit from the consensus information fused from multiple views. 

With the features from the $v$-th view, one can construct a kernel/similarity matrix for this view. There are different types of similarity matrices:
\begin{itemize}
    \item Gaussian Kernel Weighting: $W_{ij} = e^{-(\mathbf{x}_i - \mathbf{x}_j)^2/\sigma^2}$
    \item Dot-product Kernel Weighting: $W_{ij} = \mathbf{x}_i^T\cdot \mathbf{x}_j$
    \item 0-1 Weighting: $W_{ij}=1$ if and only if $\mathbf{x}_i$ is within $\mathbf{x}_j$'s k Nearest Neighbors.
\end{itemize}

A similarity matrix can then be used to define the transition probability as follows.
\begin{equation}
\mathcal{P}^{(v)}_{ij} = \frac{W_{ij}^{(v)}}{\sum_{k=1}^n W_{ik}^{(v)}}
\end{equation}
where $\sum_{j=1}^n \mathcal{P}^{(v)}_{ij} = 1$ ($\forall i = 1, \dots, n$) and we let $\mathcal{P}^{(v)}_{ii}=0$ for convenience. For a probability vector $\mathbf{u}$ (i.e., $\mathbf{u}^T\mathbf{1}= 1$), $\mathbf{u}^T\Pv^{(v)}$ is a Markov random walk of $\mathbf{u}$ w.r.t. $\Pv^{(v)}$. $\Pv^{(v)}\mathbf{u}$ can be viewed as a local averaging operation with $\mathbf{W}^{(v)}$ measuring the locality. It can also be interpreted as a generalization of Parzen window estimators to functions on the local manifold \cite{Wang12}. Both $\mathbf{u}^T\Pv^{(v)}$ and $\Pv^{(v)}\mathbf{u}$ can be viewed as a diffusion process. 

\subsection{Cross Diffusion}

Cross diffusion \cite{Wang12} aims to exploit mutual enhancement of different views inspired by co-training \cite{Blum98}. The main idea of cross diffusion is to perform random walk using the transition probability from different views in an alternating manner. In the case of $m=2$, the cross diffusion process can be defined as follows. 
\begin{align}
    \Pb^{(1)}_{t+1} & = \Pv^{(1)} \cdot \Pb^{(2)}_{t} \cdot (\Pv^{(1)})^T \\
    \Pb^{(2)}_{t+1} & = \Pv^{(2)} \cdot \Pb^{(1)}_{t} \cdot (\Pv^{(2)})^T
\end{align}
where $\Pb^{(1)}_{t}$ and $\Pb^{(2)}_{t}$ are the status matrices at the $t$-th iteration for view $1$ and view $2$, respectively. For the initial values, we set $\Pb^{(1)}_1 = \Pv^{(1)}$ and $\Pb^{(2)}_1 = \Pv^{(2)}$. Since the distances between data points are usually unreliable in high-dimensional space, it is usually preferable to use the k nearest neighbors as $\Pv^{(1)}$ and $\Pv^{(2)}$. 
Under mild conditions that $\Pv^{(1)}$ and $\Pv^{(2)}$ are irreducible and aperiodic, the convergence of this process can be proved using Perron-Frobenius Theorem \cite{Perron07}. The final status matrix can be computed as the average of status matrices from two views: $\Pb^* = (\Pb^{(1)}_{e} + \Pb^{(2)}_{e})/2$, where $e$ is the number of iterations at which the cross diffusion terminates. We refer to this final status matrix $\Pb^*$ as {\it cross diffused matrix}.

Let us denote the connected components in the cross-diffused matrix as $\{\theta_1, \theta_2, \dots, \theta_Q\}$, where $Q$ is the total number of connected components. We also denote the ground-truth class label of $\mathbf{x}$ as $c(\mathbf{x})$. We define the purity of the $q$-th connected component as the percentage of majority class of instances.
If $purity(\theta_q) \geq 1 - \epsilon$ for all $1 \leq q \leq Q$, we say that $\mathbf{P}$ is an {\bf $\epsilon$-good graph}. At the $(2t+1)$-th iteration, $\Pb^{(1)}_{2t+1}$ and $\Pb^{(2)}_{2t+1}$ can be written as the following.
\begin{align}
\Pb^{(1)}_{2t+1} & \propto (\Pv^{(1)}\Pv^{(2)})^t \cdot \Pv^{(2)} \cdot ((\Pv^{(2)})^T(\Pv^{(1)})^T)^t \\
\Pb^{(2)}_{2t+1} & \propto (\Pv^{(2)}\Pv^{(1)})^t \cdot \Pv^{(1)} \cdot ((\Pv^{(1)})^T(\Pv^{(2)})^T)^t
\end{align}

In order to effectively guide subsequent feature selection, it is desirable that the connected components in $\Pb^{(1)}_{2t+1}$ and $\Pb^{(2)}_{2t+1}$ obtained from the cross-diffusion process have large purity. The following theorem provides guarantee on the purity of components in the cross-diffused matrix \cite{Wang12}.
\begin{theorem}
If the K-nearest-neighbors is good to measure local affinity \cite{Wang10}, $\mathbf{P}^{(1)}_{2t+1}$ and $\mathbf{P}^{(2)}_{2t+1}$ are $\epsilon$-good graphs. The number of connected components in graph $\mathbf{P}^{(1)}_{2t+1}$ is equal to that of graph $\mathbf{P}^{(2)}_{2t+1}$, which is no larger than that in graphs $\mathbf{\Pv}^{(1)}$ and $\mathbf{\Pv}^{(2)}$.
\end{theorem}

Moreover, it is usually helpful to add regularization at each iteration of the diffusion process to make the probability matrix more robust. 
\begin{align}
    \Pb^{(1)}_{t+1} & = \Pv^{(1)} \cdot \Pb^{(2)}_{t} \cdot (\Pv^{(1)})^T + \alpha \mathbf{I} \\
    \Pb^{(2)}_{t+1} & = \Pv^{(2)} \cdot \Pb^{(1)}_{t} \cdot (\Pv^{(2)})^T + \alpha \mathbf{I}
\end{align}
where $\mathbf{I}$ is an identity matrix and $\alpha$ is the parameter that controls the regularization. We remark that CDMA-FS can perform reasonably well for a wide range of $\alpha$  (e.g., $10^{-4} \sim 10$).

\subsection{Extension to more than two views}

Similar to the case of $m=2$, $\Pb^{(v)}_{t+1}$ for $m>2$ can be calculated as follows.
\begin{equation}
\Pb^{(v)}_{t+1} = \Pv^{(v)} \cdot \frac{1}{m-1} \sum_{i \neq v} \Pb^{(i)}_{t} \cdot (\Pv^{(v)})^T
\end{equation}

The final status matrix is the average of $m$ matrices:
\begin{equation}
\Pb^* = \frac{1}{m} \sum_{v=1}^m \Pb^{(v)}_e
\end{equation}
Since the transition probability might be not reliable for non-nearest neighbors, we create a $k$NN graph $\mathbf{G}$ from $\Pb^*$ after obtaining $\Pb^*$. In the following section, we present how to use $\mathbf{G}$ to guide the feature selection for each view.


\section{Aligning with Cross-diffused Matrix}

Our goal is to select $d^{(v)}$ ($d^{(v)} \ll D^{(v)}$) high-quality features for each view. We denote the selection indicator vector as $\mathbf{s}^{(v)} \in \{0, 1\}^{D^{(v)}}$, where $s_p^{(v)} = 1$ indicates that the $p$-th feature is selected and $s_p^{(v)}=0$ otherwise. 

To directly exploit the information from the cross-diffused matrix for feature selection in each view, we propose to perform matrix alignment towards the cross-diffused matrix. We assume that a kernel matrix can be constructed from each view based on the selected features $\diag(\mathbf{s})\mathbf{X}^{(v)}$ with Gaussian kernels (i.e., Radial Basis Function):
\begin{equation}
K_{ij}^{(v)}=\text{exp}\left(-\frac{1}{\sigma^2}\|\text{diag}(\mathbf{s}^{(v)})\mathbf{x}_i^{(v)}-\text{diag}(\mathbf{s}^{(v)})\mathbf{x}_j^{(v)}\|^2\right)
\label{eq:rbf}
\end{equation}

The intuitive idea of CDMA-FS is to make the kernel constructed from selected features imitate the cross-diffused matrix $\mathbf{G}$. We achieve this by employing the matrix alignment technique \cite{Cristianini01} \cite{Wei16c} as follows.
\begin{mydef}
{\bf Matrix Alignment} For two symmetric matrices $\K_1 \in \mathbb{R}^{n\times n}$ and $\K_2 \in \mathbb{R}^{n\times n}$, the alignment between $\K_1$ and $\K_2$ is defined as 
\begin{equation}
\rho(\K_1, \K_2) = \frac{\Tr(\K_1\K_2)}{||\K_1||_F \cdot ||\K_2||_F}
\end{equation}
where $\Tr(\cdot)$ is the trace of a matrix.
\end{mydef}

Matrix alignment can be viewed as calculating the cosine similarity between two vectorized matrices. However, the normalization term $||\K_1||_F \cdot ||\K_2||_F$ makes the optimization problem more difficult to solve. In this paper, we employ the unnormalized version of matrix alignment as in \cite{Cristianini01}, which can be considered as the inner product between two vectorized matrices.
\begin{mydef}
{\bf Unnormalized Matrix Alignment} For two symmetric matrices $\K_1 \in \mathbb{R}^{n\times n}$ and $\K_2 \in \mathbb{R}^{n\times n}$, the alignment between $\K_1$ and $\K_2$ is defined as 
\begin{equation}
\rho(\K_1, \K_2) = \Tr(\K_1  \K_2)
\end{equation}
\end{mydef}
It is usually helpful to center the matrix for better matrix alignment performance as in observed in \cite{Cortes12}. For a symmetric matrix $\mathbf{K}$, centering $\mathbf{K}$ can be achieved by $\mathbf{HKH}$, where the centering matrix $\mathbf{H} = \mathbf{I}-\frac{1}{n}\mathbf{1}\mathbf{1}^T$.
\begin{mydef}
{\bf Centered Matrix Alignment} For two real matrices $\K_1 \in \mathbb{R}^{n\times n}$ and $\K_2 \in \mathbb{R}^{n\times n}$, the centered alignment between $\K_1$ and $\K_2$ is defined as 
\begin{align*}
\rho(\K_1, \K_2) = & \Tr(\mathbf{H}\K_1\mathbf{H}\mathbf{H}\K_2\mathbf{H}) \\
                 = & \Tr(\mathbf{H}\K_1\mathbf{H}\K_2)
\end{align*}
where the second equation can be obtained by noting $\mathbf{H}\mathbf{H} = \mathbf{H}$ and $\Tr(\mathbf{AB})=\Tr(\mathbf{BA})$ for arbitrary matrices $\mathbf{A}, \mathbf{B} \in \mathbb{R}^{n\times n}$.
\end{mydef}

After a high-quality cross-diffused matrix is obtained, we select features for each view under the guidance of this matrix. To achieve this, we aim to maximize the correlation between the cross-diffused matrix and the kernel matrix computed from selected features. To select $d^{(v)}$ features for the $v$-th view, we formulate it as a constrained optimization problem and find $\mathbf{s}^{(v)}$ to minimize the following objective function:
\begin{align}
\begin{split}
\operatornamewithlimits{min}_{\mathbf{s}^{(v)}} ~ & f= - \Tr(\mathbf{HGHK}^{(v)}) \\
\text{s.t.} ~ & \sum_{p=1}^{D^{(v)}} s_p^{(v)} = d^{(v)} \\
  ~ & s_p^{(v)} \in \{0, 1\}, \forall p = 1, \dots, D^{(v)}
\label{eq:obj}
\end{split}
\end{align}

\noindent {\bf Discussion} Traditional sparse regression based methods \cite{Tang13} \cite{Qian14} rely on generating intermediate cluster labels and rank features by their linear regression coefficients. In contrast, CDMA-FS framework utilizes the cross-diffused matrix, which preserves more information than cluster labels.
Also, the connected components in the cross diffused matrix tend to have good purity as shown in Theorem 1, which means the connected data points are likely from the same class. The objective, through matrix alignment, aims to select the features that make connected instances close and unconnected instances far apart. By optimizing the objective above, we directly infer the selection vector $\mathbf{s}$ which can achieve the following desirable effects: features that make data points from the same class similar would be rewarded and features that make data points from different classes similar would shrink $s_p$ to zero. Hence, different classes would be more separable in the space of selected features.


\section{Optimization}
\subsection{Gradient Derivation with Relaxed Constraint}

The `0/1' integer programming problem in Eq (\ref{eq:obj}) is computationally intensive to optimize. We relax the `0/1' constraint on $s^{(v)}_p$ ($p = 1, \dots, D^{(v)}$) to real values in range of $[0, 1]$  to make the optimization tractable as in \cite{Wei16}. We further rewrite the summation constraint $\sum_{p=1}^{D^{(v)}} s_p^{(v)} = d^{(v)}$ in the form of Lagrange multiplier:
\begin{align}
\begin{split}
\operatornamewithlimits{min}_{\mathbf{s}^{(v)}} ~ & f= - \Tr(\mathbf{HGHK}^{(v)}) + \lambda ||\mathbf{s}^{(v)}||_1 \\
\text{s.t.}  ~ & 0 \leq s_p^{(v)} \leq 1, \forall p = 1, \dots, D^{(v)}
\end{split}
\label{eq:objrelax}
\end{align}
where $||\cdot||_1$ denotes the $l_1$ norm on vector $(\cdot)$ and $\lambda$  controls the sparsity of $\mathbf{s}^{(v)}$. Note that in our case $||\mathbf{s}^{(v)}||_1 = \sum_{p=1}^D s_p$ since we have non-negative constraints on $\mathbf{s}^{(v)}$.


We can derive the following gradient w.r.t. the objective function, since $\mathbf{K}^{(v)}$ ($v = 1, \dots, m$) is a symmetric matrix.
\begin{align}
\begin{split}
\frac{\partial f}{\partial s_p^{(v)}} & = -\sum_{i,j=1}^n ((\mathbf{HGH})_{ij} \cdot \frac{\partial K_{ij}^{(v)}}{\partial s_p^{(v)}}) + \lambda \\
   & = \sum_{i,j=1}^n (((\mathbf{HGH}) \odot \mathbf{K}^{(v)})_{ij} \left(x_{ip}^{(v)}-x_{jp}^{(v)}\right)^2) \frac{2 s_p}{\sigma^2} + \lambda 
\end{split}
   \label{eq:gradient}
\end{align}
where $\odot$ is element-wise product. To solve this constrained optimization problem efficiently, we use Projected Quasi-Newton Method as shown in the next subsection.

\subsection{Projected Quasi-Newton Method}

Traditional Newton method optimizes the following second-order approximation at the $t$-th iteration.
\begin{equation}
q_t(\mathbf{s}) = f(\mathbf{s}_t) + (\mathbf{s}-\mathbf{s}_t)^T\nabla f(\mathbf{s}_t) + \frac{1}{2}(\mathbf{s}-\mathbf{s}_t)^T \mathbf{B}_t(\mathbf{s}-\mathbf{s}_t)
\end{equation}
where $\mathbf{B}_t=\nabla^2 f(\mathbf{s}_t)$ is the Hessian matrix. Newton method enjoys good convergence rate but the Hessian matrix requires $O(D^2)$ storage and it is time-consuming to compute. So Quasi-Newton methods (e.g., L-BFGS \cite{Liu89}) use a positive definite approximation to the Hessian matrix $\nabla^2 f(\mathbf{s}_t)$. For example, L-BFGS \cite{Liu89} uses the gradients in previous iterations to compute an approximate Hessian matrix. 
\begin{equation}
\mathbf{B}_{t+1} = \mathbf{B}_t - \frac{\mathbf{B}_t \mathbf{u}_t \mathbf{u}_t^T \mathbf{B}_t}{\mathbf{u}_t^T \mathbf{B}_t \mathbf{u}_t} + \frac{\mathbf{y}_t \mathbf{y}_t^T}{\mathbf{y}_t^T \mathbf{u}_t}
\label{eq:bfgs}
\end{equation}
where $\mathbf{u}_t = \mathbf{s}_{t+1} - \mathbf{s}_t$ and $\mathbf{y}_t = \nabla f(\mathbf{s}_{t+1}) - \nabla f(\mathbf{s}_t)$.

To address the constraints on $\mathbf{s}$ in Eq (\ref{eq:objrelax}), projected Newton method can be used to solve the following constrained quadratic approximation:
\begin{align}
\begin{split}
\min_{\mathbf{s}} ~ & ~ q_t(\mathbf{s}) \\
  \text{s.t.} ~ &  ~ \mathbf{s} \in \mathcal{C}
  \label{eq:approxi}
\end{split}
\end{align}
In our case, $\mathcal{C}$ is the $[0, 1]$ box constraint on $\mathbf{s}^{(v)}$. A projection operator for this constraint can be defined as follows.
\begin{equation}
[\text{Proj}_{[0, 1]}(\mathbf{s}^{(v)})]_p = \text{min}(1, \text{max}(0, s_p^{(v)})), ~ \forall p = 1, 2, \dots, D^{(v)}
\end{equation}
To make the optimization more efficient, we use a variant of the L-BFGS method which employs spectral projected gradient method as subroutine to solve the constrained problem in Eq (\ref{eq:approxi}). The optimization method \cite{Schmidt09} is two-level approach: at the outer level, L-BFGS updates are used to construct a sequence of quadratic approximations (with constraints) to the problem; at the inner level, a spectral projected gradient method optimizes the constrained subproblem approximately to generate a feasible direction. The number of iterations in this algorithm remains linear in dimensionality of feature vector, but with a higher constant factor than the L-BFGS method. Nevertheless, the method can lead to significant gain when the cost of the projection is much lower than evaluating the function, which is the case in our problem setting.

Although we could use spectral projected gradient method to exactly solve problem Eq (\ref{eq:approxi}), it is expensive to do so in practice. Therefore, we terminate the spectral gradient descent subroutine before the exact solution is found, since our goal is only to obtain a feasible descent direction for L-BFGS. One might be concerned about the early termination of the spectral gradient descent subroutine, but in \cite{Schmidt09} it has been shown that the spectral gradient descent subroutine, even when terminated early, can give a descent direction, if we initialize it with $\mathbf{s}_t$ and we perform at least one spectral gradient descent iteration. In the implementation, we can parametrize the maximum number of the spectral gradient descent iterations by $t_p$, the cost of one iteration is $O(mt_pD)$ for the inexact Newton method, given that our projection operation requires $O(D)$ time and L-BFGS stores $m$ most recent gradients. The projected Quasi-Newton algorithm is shown in Algorithm \ref{alg:pqn}.


\begin{algorithm}[ht]
  \caption{Solve CDMA-FS with Projected Quasi-Newton Algorithm}

  \begin{algorithmic}
  \State Initialize: $ \mathbf{s}_0 \leftarrow \mathbf{1}$, $t=0$.
  \While {not converged}
  \State Compute the gradient by Eq (\ref{eq:gradient})
  \State Compute the approximate Hessian
  \State Solve Eq (\ref{eq:approxi}) for $\mathbf{s}_t^*$ using projected spectral gradient algorithm. 
  \State $\mathbf{d}_t = \mathbf{s}_t^* - \mathbf{s}_t$ 
  \State Perform line search on the direction of $\mathbf{d}_t$ to satisfy the Armijo condition.
  \State $t=t+1$
  \EndWhile
  \State Select the features with corresponding entry in $\mathbf{s}$ equal to $1$.
  \end{algorithmic}
  \label{alg:pqn}
\end{algorithm}

\section{Parameter Selection}
Existing multi-view feature selection methods typically have $2 \sim 3$ regularization parameters and it is difficult to choose appropriate values for these parameters when class labels are not available. In the original papers of these psuedo-label approaches \cite{Yang11} \cite{Qian13} \cite{Shi14}, only the best performance is reported, the parameters of which are tuned using all the class labels. However, such way of setting parameters violates the assumption of no supervision. In practice, it is impossible to know the best parameter values and this makes them less useful for real world applications. 

For CDMA-FS, we provide guidelines for choosing the value of parameter $\lambda$. Let us denote the number of features with $s_p^{(v)}=1$ as $N_1^{(v)}$, which is influenced by the value of $\lambda$. By noting that $N_1^{(v)}$ is a monotonically non-increasing function of $\lambda$, we can choose the value of $\lambda$ for each view that makes $N_1^{(v)}$ equal to (or within a small range of) the feature size one wants to retain.

\begin{table*}[htbp]
\begin{center}
    \caption{Statistics of datasets}
    \begin{tabular}{| p{2.4cm} || l | l | l | l | l | l | l | l|}
    \hline
    Statistics & \multicolumn{2}{l|}{Reuters} & \multicolumn{2}{l|}{BBC Sport} & \multicolumn{2}{l|}{BlogCatalog} &  \multicolumn{2}{l|}{CNN} \\ \hline
    \# of instances & \multicolumn{2}{l|}{1575} & \multicolumn{2}{l|}{544} & \multicolumn{2}{l|}{1000} &  \multicolumn{2}{l|}{2107}    \\ \hline
    \multirow{2}{*}{\# of features} & view1 & view2 & view1 & view2 & view1 & view2 & view1 & view2 \\ 
    & 3791 & 2862 & 3183 & 3203 & 5390 & 2003 & 6262 & 996  \\ \hline
    \# of classes & \multicolumn{2}{l|}{6} & \multicolumn{2}{l|}{5} &  \multicolumn{2}{l|}{5}  & \multicolumn{2}{l|}{7}   \\
    \hline
    \end{tabular}
    \label{table:dataset}
\end{center}
\end{table*}

\section{Experiments}

In this section, we compare the proposed method with state-of-the-art baseline methods on four real world datasets.

\subsection{Datasets}
We use four publicly available real-world datasets in our experiments.
\begin{itemize}
    \item Reuters Multilingual dataset \footnote{\url{https://archive.ics.uci.edu/ml/datasets/Reuters+RCV1+RCV2+Multilingual,+Multiview+Text+Categorization+Test+collection}}: News articles in English and German on six topics. Each language can be considered a view for the same article.
    \item BBC Sport dataset \footnote{\url{http://mlg.ucd.ie/datasets/segment.html}}: BBC news articles from $5$ topics: {\it athletics, cricket, football, rugby, tennis}. Paragraphs in the news articles are used to construct two views.
    \item CNN dataset \footnote{\url{https://sites.google.com/site/qianmingjie/home/datasets/cnn-and-fox-news}}: It consists of news articles from CNN with two views: news text and images in the news.
    \item Blogcatalog dataset \footnote{\url{http://dmml.asu.edu/users/xufei/datasets.html}}: A subset of blog posts from Blogcatalog website in the categories of \{Autos, Software, Crafts, Football, Career\&Jobs\}. Two views are the text in posts and the tags associated with the posts, respectively.
\end{itemize}


The statistics of four real-world datasets is summarized in Table \ref{table:dataset}.

\begin{table*}[!htbp]
\begin{center}
\caption{Clustering accuracy on four datasets. For the baselines that need parameter tuning, best/median performance is reported. }
   \scalebox{0.9}{
    \begin{tabular}{| l || l | l | l | l || l | l | l | l |}
    \hline
    Method & \multicolumn{4}{c||}{{\bf BBC Sport}} & \multicolumn{4}{c|}{{\bf Reuters}}  \\ \hline
    \# features   & 100 & 200 & 300 & 400 & 100 & 200 & 300 & 400  \\ \hline
    All Features & \multicolumn{4}{c||}{0.5960} & \multicolumn{4}{c|}{0.6545}  \\ \hline
    LS &   0.4034 & 0.3885 & 0.3756 & 0.4112    &  0.3792 & 0.4587 & 0.5446 & 0.5900     \\ \hline
    UDFS &  0.4565/0.4504 & 0.5232/0.5228 & 0.5549/0.5107 &  0.5525/0.5164    &   0.4320/0.4225 & 0.4921/0.4436 & 0.5926/0.4630 & 0.5918/0.5421     \\ \hline
    RSFS &  0.6054/0.5388 & 0.6515/0.5709 & 0.6713/0.6041 & 0.6634/0.6085        &    0.5688/0.4558 & 0.5757/0.4529 & 0.6546/0.5271 & 0.6259/0.5332      \\ \hline
    MVFS &  0.5996/0.5480 & 0.6572/0.5662 & 0.6148/0.5966 & 0.6118/0.6015    &   0.5302/0.4284 & 0.5561/0.4505 & 0.5592/0.5447 & 0.5950/0.5299     \\ \hline
    MVUFS & 0.6253/0.4338 & 0.6181/0.5258 & 0.6242/0.6089 & 0.6542/0.6181     &   0.5998/0.3677 & 0.6476/0.4782 & 0.6397/0.5339 & 0.6182/0.5619     \\ \hline
    CDMA-FS & 0.7341 & 0.7403 & 0.7472 & 0.7494     &  0.5465 & 0.6015 & 0.6322 & 0.6428   \\ \hline
    \hline
    Method & \multicolumn{4}{c||}{{\bf BlogCatalog}} & \multicolumn{4}{c|}{{\bf CNN}} \\ \hline
    \# features   & 100 & 200 & 300 & 400 & 100 & 200 & 300 & 400  \\ \hline 
    All Features & \multicolumn{4}{c||}{0.5979} & \multicolumn{4}{c|}{0.3005} \\ \hline
    LS &    0.3947 & 0.3975 & 0.4112 & 0.4550    &  0.2435 & 0.2419 & 0.2573 & 0.3238     \\ \hline
    UDFS &   0.5219/0.4153 & 0.6173/0.6022 & 0.6561/0.6556 & 0.6489/0.6459 
   &   0.4095/0.4084 & 0.4019/0.3956 & 0.4171/0.3921 & 0.3962/0.3772   \\ \hline
    RSFS &   0.6388/0.4995 & 0.6504/0.5733 & 0.6657/0.5917 & 0.6513/0.6014    &  0.3647/0.2692 & 0.4131/0.3140 & 0.4112/0.3608 & 0.4243/0.3596    \\ \hline
    MVFS &   0.5409/0.5139 & 0.6027/0.5690 & 0.6107/0.5778 & 0.6457/0.6056   &  0.3578/0.2639 & 0.4204/0.3511 & 0.3902/0.3697 & 0.4213/0.3637   \\ \hline
    MVUFS &    0.6157/0.4901 & 0.6693/0.6157 & 0.6565/0.5514 & 0.6496/0.5521   &   0.4524/0.3227 & 0.4899/0.3520 & 0.4879/0.3402 & 0.4649/0.3566   \\ \hline
    CDMA-FS  &   0.6029 & 0.6746 & 0.6704 & 0.6851    &   0.5347 & 0.4989 & 0.4771 & 0.4783 
   \\ \hline
    \hline
    \end{tabular}
    }
    \label{table:acc}
\end{center}
\end{table*}

\begin{table*}[htbp]
\begin{center}
\caption{Clustering NMI on four datasets. For the baselines that need parameter tuning, best/median performance is reported. }
   \scalebox{0.9}{
    \begin{tabular}{| l || l | l | l | l || l | l | l | l |}
    \hline
    Method & \multicolumn{4}{c||}{{\bf BBC Sport}} & \multicolumn{4}{c|}{{\bf Reuters}}  \\ \hline
    \# features   & 100 & 200 & 300 & 400 & 100 & 200 & 300 & 400  \\ \hline
    All Features & \multicolumn{4}{c||}{0.4434 } & \multicolumn{4}{c|}{0.4846}  \\ \hline
    LS &  0.0724 & 0.0775 & 0.0702 & 0.1099      &   0.1960 & 0.2689 & 0.3486 & 0.3989    \\ \hline
    UDFS & 0.2279/0.1968 & 0.3453/0.2994 & 0.3453/0.2939 & 0.3386/0.2861   &  0.2203/0.2187 & 0.2829/0.2639 & 0.4023/0.2834 & 0.4046/0.3677  \\ \hline
    RSFS &  0.3543/0.3141 & 0.4340/0.3900 & 0.5162/0.4151 & 0.5076/0.4166        &  0.4079/0.2429 & 0.4329/0.2963 & 0.4539/0.3648 & 0.4666/0.4134    \\ \hline
    MVFS &  0.3383/0.3133 & 0.4288/0.3899 & 0.4276/0.4155 & 0.4371/0.4157    &  0.3594/0.2267 & 0.3986/0.2787 & 0.4256/0.3855 & 0.4427/0.4180     \\ \hline
    MVUFS &  0.4374/0.2062 & 0.4255/0.3171 & 0.4273/0.4032 & 0.4443/0.4236    &  0.4260/0.1866 & 0.4887/0.3346 & 0.4816/0.3570 & 0.4681/0.4000    \\ \hline
    CDMA-FS &  0.5774 & 0.6659 & 0.6693 & 0.6738   & 0.3823 & 0.4532 & 0.4801 & 0.4858     \\ \hline
    \hline
    Method & \multicolumn{4}{c||}{{\bf BlogCatalog}} & \multicolumn{4}{c|}{{\bf CNN}} \\ \hline
    \# features   & 100 & 200 & 300 & 400 & 100 & 200 & 300 & 400  \\ \hline 
    All Features & \multicolumn{4}{c||}{0.4782} & \multicolumn{4}{c|}{0.0957} \\ \hline
    LS &  0.2252 & 0.2458 & 0.2400 & 0.2819      &  0.0513 & 0.0557 & 0.0667 & 0.1280     \\ \hline
    UDFS & 0.3223/0.1978 & 0.4123/0.3580 & 0.4501/0.4309 & 0.4753/0.4328   &  0.2122/0.1897 & 0.1852/0.1846 & 0.1920/0.1831 & 0.1868/0.1784  \\ \hline
    RSFS &  0.4260/0.3090 & 0.4551/0.3564 & 0.4715/0.4064 & 0.4746/0.4408       &  0.1537/0.0690 & 0.1862/0.0984 & 0.1853/0.1430 & 0.2048/0.1383    \\ \hline
    MVFS &  0.3432/0.3181 & 0.3971/0.3543 & 0.4274/0.4041 & 0.4764/0.4424    &  0.1517/0.0739 & 0.2051/0.1391 & 0.1558/0.1444 & 0.2034/0.1391     \\ \hline
    MVUFS & 0.4237/0.2910 & 0.4747/0.4347 & 0.4643/0.3997 & 0.4504/0.3998     &  0.2242/0.1170 & 0.2824/0.1340 & 0.2917/0.1423 & 0.2670/0.1645    \\ \hline
    CDMA-FS &  0.4176 & 0.4650 & 0.4866 & 0.5105   &   0.3244 & 0.3244 & 0.3049 & 0.2910   \\ \hline
    \hline
    \end{tabular}
    }
    \label{table:nmi}
\end{center}
\end{table*}

\subsection{Baselines}

We compare CDMA-FS with using all features and five other unsupervised feature selection methods as follows:

\begin{itemize}
\item All Features: It uses all original features without selection for evaluation.
\item LS: Laplacian Score \cite{He05} selects the features that preserve the local manifold structure.
\item UDFS: Unsupervised Discriminative Feature Selection \cite{Yang11} is a pseudo-label based approach with $L_{2,1}$ regularization to exploit the local structure. 
\item RSFS: Robust Spectral Feature Selection \cite{Shi14} selects features by robust spectral analysis framework with sparse regression.
\item MVFS: Multi-view Feature Selection \cite{Tang13} is unsupervised feature selection for multi-view data based on pseudo labels, which are generated as the consensus of spectral clustering on two views.
\item MVUFS: Multi-view Unsupervised Feature Selection \cite{Qian14} generates pseudo-labels by Non-negative Matrix Factorization and local kernel learning.
\end{itemize}

\subsection{Experiment setup}

In this section, we evaluate the quality of selected features by their clustering performance. We use the the popular co-regularized spectral clustering \cite{Kumar11} for clustering multi-view data \footnote{We use the code at \url{http://www.umiacs.umd.edu/~abhishek/code_coregspectral.zip}}. We set their $\sigma$ as the median of pairwise Euclidean distances between data points and $\lambda =0.1$ as suggested in the paper. KMeans is then used on these latent factors. We repeat the KMeans experiment for 20 times (since it is initilization) and report the average performance. We vary the number of features $d$ in the range of \{$100$, $200$, $300$, $400$\}. For each feature size $d$, we choose appropriate $\lambda$ in our method via binary search to let the number of selected features (with score $s_p=1$) within $d\pm 10$.

\begin{figure*}[t]
\centering
    \begin{subfigure}[b]{0.41\textwidth}
        \centering
        \includegraphics[height=2.3in]{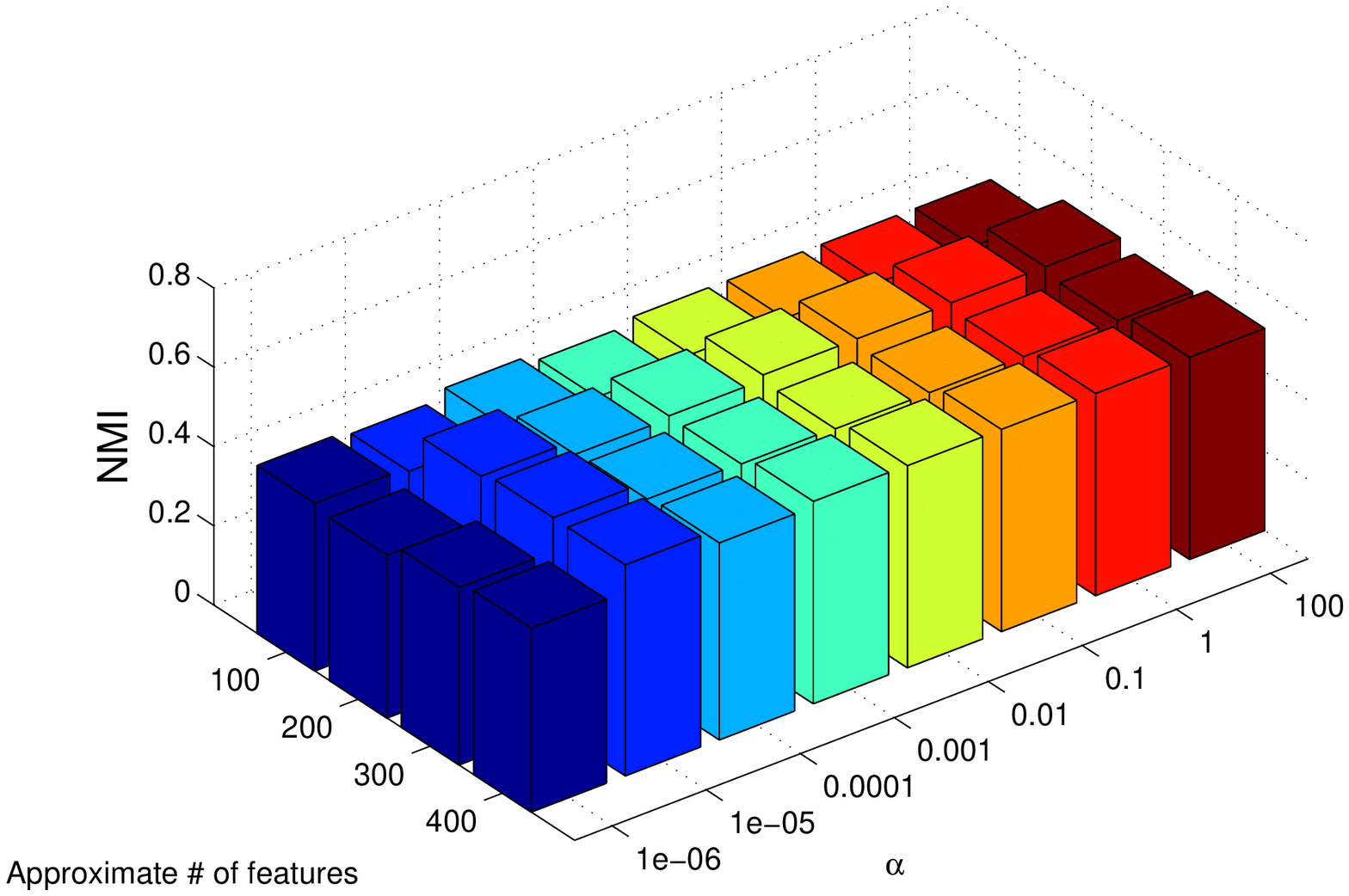}
        \caption{BlogCatalog}
    \end{subfigure}%
\begin{subfigure}[b]{0.41\textwidth}
        \centering
        \includegraphics[height=2.3in]{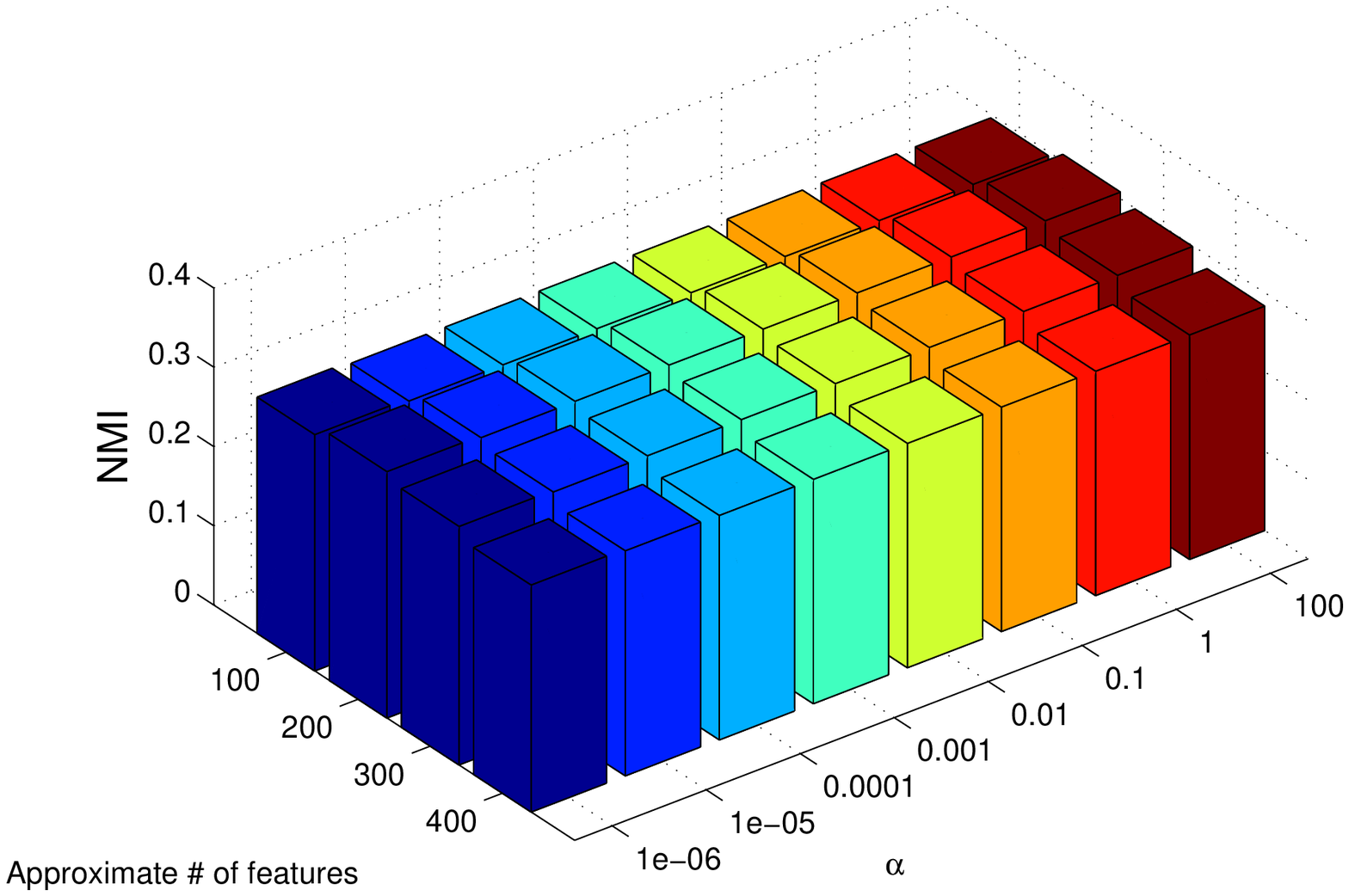}
        \caption{CNN}
    \end{subfigure}
\caption{NMI w.r.t different values of $\alpha$}
\label{fig:sensitivity}
\end{figure*}

Following the typical experimental setting for unsupervised feature selection \cite{Yang11} \cite{Li12} \cite{Wei16a}, we use Accuracy and Normalized Mutual Information (NMI) to evaluate the result of clustering. Accuracy is defined as follows. 
\begin{equation}
Accuracy = \frac{1}{n} \sum_{i=1}^n \mathcal{I}(c_i= map(p_i))
\end{equation}
where $p_i$ is the clustering result of instance $i$ and $c_i$ is its real class label. $map(\cdot)$ is a mapping function that maps each cluster label to a ground-truth label using Kuhn-Munkres Algorithm \cite{Munkres57}.

Normalized Mutual Information (NMI) is another popular metric for evaluating clustering performance. Let $C$ be the set of clusters from the ground truth and $C'$ obtained from a clustering algorithm. Their mutual information $MI(C, C')$ can be defined as follows:
\begin{equation}
MI(C, C') = \sum_{c_i \in C, c'_j \in C'} p(c_i, c'_j) \log \frac{p(c_i, c'_j)}{p(c_i)p(c'_j)}
\end{equation}
where $p(c_i)$ and $p(c'_j)$ are the probabilities that a random instance from the data set belongs to $c_i$ and $c'_j$, respectively, and $p(c_i, c'_j)$ is the joint probability that the instance belongs to the cluster $c_i$ and $c'_j$ at the same time. In our experiments, we use the normalized mutual information as in previous work \cite{Li12} \cite{Shi14}. 
\begin{equation}
NMI(C, C') = \frac{MI(C, C')}{max(H(C), H(C'))}
\end{equation}
where $H(C)$ and $H(C')$ are the entropy of $C$ and $C'$. Higher value of and Accuracy and NMI indicates better quality of clustering.

We set $k=5$ for the kNN neighbor size in the baseline methods and our approach following previous convention \cite{Li12}. For the number of pseudo-classes in UDFS, RSFS, MVFS and MVUFS, we use the ground-truth number of classes.  Also, we perform grid search in the range of $\{0.1, 1, 10\}$ for the regularization parameters in these baseline methods. Besides their best performance, we also report the median performance for them. For CDMA-FS proposed in this paper, we use `0/1' weighting in the $\mathbf{W}$ and we fix $\sigma^2 = 1$ and $\alpha=0.01$ for all the datasets after normalizing each data point to unit length. We set the maximum number of iterations for the cross-diffusion process as $20$.




\subsection{Results}

The clustering accuracy and NMI on four datasets are shown in Table \ref{table:acc} and \ref{table:nmi}. It can be observed that feature selection is a useful technique for improving the multi-view clustering performance. For example, compared with using all the features, CDMA-FS with $400$ features improves the accuracy on BBCSport and BlogCatalog datasets by $26\%$ and $15\%$, respectively.
When comparing with other feature selection methods, we can observe that CDMA-FS performs favorably or comparable to the best performance of baseline methods, the parameters of which are tuned using all the class labels. Considering that in practice one cannot know the best parameters for these baseline methods (since we assume no supervision), their median performance is a better reflection of these methods' practical power, which is far inferior to CDMA-FS.

\subsection{Parameter Sensitivity}
In this subsection, we study how the regularization $\alpha$ in the cross diffusion process affects the quality of selected features. The performance w.r.t different $\alpha$ on BlogCatalog and CNN is shown in Figure~\ref{fig:sensitivity}. We can observe that the performance is not very sensitive to $\alpha$, and CDMA-FS can perform reasonably well when $\alpha>10^{-5}$. In contrast, the baseline methods in Table \ref{table:acc} and \ref{table:nmi} tend to be more sensitive w.r.t. the parameter values, as the their median performance differs significantly with best performance.

\section{Conclusion}

High-dimensional multi-view data pose challenges for many machine learning tasks. While feature selection methods can be useful for alleviating the curse of dimensionality, existing approaches either cannot exploit information from multiple views simultaneously or rely on cluster labels for this task. In this paper, we aim to preserve more accurate information from multi-view data by learning a cross-diffused matrix and directly utilize the information by matrix alignment. Experimental results show that CDMA-FS is able to select high-quality features on real-world datasets and outperforms the baseline methods significantly.

\bibliographystyle{abbrv}
\bibliography{reference}

\begin{thebibliography}{10}

\bibitem{Blum98}
A.~Blum and T.~Mitchell.
\newblock Combining labeled and unlabeled data with co-training.
\newblock pages 92--100, 1998.

\bibitem{Cortes12}
C.~Cortes, M.~Mohri, and A.~Rostamizadeh.
\newblock Algorithms for learning kernels based on centered alignment.
\newblock {\em Journal of Machine Learning Research}, 13:795--828, 2012.

\bibitem{Cristianini01}
N.~Cristianini, J.~Shawe-Taylor, A.~Elisseeff, and J.~S. Kandola.
\newblock On kernel-target alignment.
\newblock In {\em NIPS}, pages 367--373, 2001.

\bibitem{Du15}
L.~Du and Y.-D. Shen.
\newblock Unsupervised feature selection with adaptive structure learning.
\newblock In {\em KDD}, pages 209--218, 2015.

\bibitem{Feng12}
Y.~Feng, J.~Xiao, Y.~Zhuang, and X.~L. 0002.
\newblock Adaptive unsupervised multi-view feature selection for visual concept
  recognition.
\newblock In {\em ACCV (1)}, volume 7724, pages 343--357, 2012.

\bibitem{He05}
X.~He, D.~Cai, and P.~Niyogi.
\newblock Laplacian score for feature selection.
\newblock In {\em NIPS}, 2005.

\bibitem{Kumar11}
A.~Kumar, P.~Rai, and H.~D. III.
\newblock Co-regularized multi-view spectral clustering.
\newblock In {\em NIPS}, pages 1413--1421, 2011.

\bibitem{Li12}
Z.~Li, Y.~Yang, J.~Liu, X.~Zhou, and H.~Lu.
\newblock Unsupervised feature selection using nonnegative spectral analysis.
\newblock In {\em AAAI}, 2012.

\bibitem{Liu89}
D.~C. Liu and J.~Nocedal.
\newblock On the limited memory bfgs method for large scale optimization.
\newblock {\em Mathematical Programming}, 45:503--528, 1989.

\bibitem{Munkres57}
J.~Munkres.
\newblock Algorithms for the assignment and transportation problems.
\newblock {\em Journal of the Society of Industrial and Applied Mathematics},
  5(1):32--38, 1957.

\bibitem{Perron07}
O.~Perron.
\newblock {Zur Theorie der Matrices}.
\newblock {\em Mathematische Annalen}, 64(2):248--263, 1907.

\bibitem{Qian13}
M.~Qian and C.~Zhai.
\newblock Robust unsupervised feature selection.
\newblock In {\em IJCAI}, 2013.

\bibitem{Qian14}
M.~Qian and C.~Zhai.
\newblock Unsupervised feature selection for multi-view clustering on
  text-image web news data.
\newblock In {\em CIKM}, pages 1963--1966. ACM, 2014.

\bibitem{Schmidt09}
M.~Schmidt, E.~V.~D. Berg, M.~P. Friedl, and K.~Murphy.
\newblock Optimizing costly functions with simple constraints: A limited-memory
  projected quasi-newton algorithm.
\newblock In {\em In AI \& Statistics}, 2009.

\bibitem{Shao16}
W.~Shao, L.~He, C.-T. Lu, X.~Wei, and P.~S. Yu.
\newblock Online unsupervised multi-view feature selection.
\newblock In {\em ICDM}, 2016.

\bibitem{Shi14}
L.~Shi, L.~Du, and Y.-D. Shen.
\newblock Robust spectral learning for unsupervised feature selection.
\newblock In {\em ICDM}, 2014.

\bibitem{Tang13}
J.~Tang, X.~Hu, H.~Gao, and H.~Liu.
\newblock Unsupervised feature selection for multi-view data in social media.
\newblock In {\em SDM}, pages 270--278, 2013.

\bibitem{Wang12}
B.~Wang, J.~Jiang, W.~Wang, Z.-H. Zhou, and Z.~Tu.
\newblock Unsupervised metric fusion by cross diffusion.
\newblock In {\em CVPR}, pages 2997--3004, 2012.

\bibitem{Wang15}
S.~Wang, J.~Tang, and H.~Liu.
\newblock Embedded unsupervised feature selection.
\newblock In {\em Proceedings of the Twenty-Ninth {AAAI} Conference on
  Artificial Intelligence, January 25-30, 2015, Austin, Texas, {USA.}}, pages
  470--476, 2015.

\bibitem{Wang10}
W.~Wang and Z.-H. Zhou.
\newblock A new analysis of co-training.
\newblock In {\em ICML}, pages 1135--1142, 2010.

\bibitem{Wei16c}
X.~Wei, B.~Cao, W.~Shao, C.-T. Lu, and P.~S. Yu.
\newblock Community detection with partially observable links and node
  attributes.
\newblock In {\em IEEE International Conference on Big Data}, 2016.

\bibitem{Wei16}
X.~Wei, B.~Cao, and P.~S. Yu.
\newblock Nonlinear joint unsupervised feature selection.
\newblock In {\em SDM}, 2016.

\bibitem{Wei16b}
X.~Wei, B.~Cao, and P.~S. Yu.
\newblock Unsupervised feature selection on networks: A generative view.
\newblock In {\em AAAI}, 2016.

\bibitem{Wei15}
X.~Wei, S.~Xie, and P.~S. Yu.
\newblock Efficient partial order preserving unsupervised feature selection on
  networks.
\newblock In {\em SDM}, pages 82--90. SIAM, 2015.

\bibitem{Wei16a}
X.~Wei and P.~S. Yu.
\newblock Unsupervised feature selection by preserving stochastic neighbors.
\newblock In {\em AISTATS}, 2016.

\bibitem{Yang11}
Y.~Yang, H.~T. Shen, Z.~Ma, Z.~Huang, and X.~Zhou.
\newblock l2, 1-norm regularized discriminative feature selection for
  unsupervised learning.
\newblock In {\em IJCAI}, pages 1589--1594, 2011.

\bibitem{Zhao07}
Z.~Zhao and H.~Liu.
\newblock Spectral feature selection for supervised and unsupervised learning.
\newblock In {\em ICML}, volume 227, pages 1151--1157, 2007.

\end{thebibliography}

\end{document}